\documentclass[twocolumn,showpacs,%
  nofootinbib,aps,superscriptaddress,%
  eqsecnum,prd,notitlepage,showkeys,10pt]{revtex4-1}

\usepackage{amssymb}
\usepackage{amsmath}
\usepackage{graphicx}
\usepackage{dcolumn}
\usepackage[colorlinks=true, urlcolor=blue, linkcolor=red]{hyperref}
\usepackage{float}

\begin{document}

\title{Benchmarking LLM powered Chatbots: Methods and Metrics}
\author{
{Debarag Banerjee,}
{Pooja Singh,}
{Arjun Avadhanam,}
{Saksham Srivastava}
}
\providecommand{\keywords}[1]{\textbf{\textit{Index terms---}} #1}
\begin{abstract}
\begin{center}
    \textbf{Abstract}\\
\end{center}
Autonomous conversational agents, i.e. chatbots, are becoming an increasingly common mechanism for enterprises to provide support to customers and partners.  In order to rate chatbots, especially ones powered by Generative AI tools like Large Language Models (LLMs) we need to be able to accurately assess their performance. This is where chatbot benchmarking becomes important. In this paper, we propose the use of a novel benchmark that we call the E2E (End to End) benchmark, and show how the E2E benchmark can be used to evaluate accuracy and usefulness of the answers provided by chatbots, especially ones powered by LLMs. We evaluate an example chatbot at different levels of sophistication based on both our E2E benchmark, as well as other available metrics commonly used in the state of art, and observe that the proposed benchmark show better results compared to others. In addition, while some metrics proved to be unpredictable, the metric associated with the E2E benchmark, which uses cosine similarity performed well in evaluating chatbots. The performance of our best models shows that there are several benefits of using the cosine similarity score as a metric in the E2E benchmark.
\end{abstract}

\maketitle

\section{Introduction}

“A chatbot is a computer program that uses \href{https://www.ibm.com/topics/artificial-intelligence}{artificial intelligence} (AI) and \href{https://www.ibm.com/topics/natural-language-processing}{natural language processing} (NLP) to understand customer questions and automate responses to them, simulating human conversation. – (\href{https://www.ibm.com/topics/chatbots}{"What is a chatbot?"}) As the field of AI continues to grow rapidly, chatbots have become more and more refined, now capable of learning from large volumes of data and improving their conversational abilities over time. This has given rise to a new section of chatbots known as ML (Machine Learning) based chatbots that are not pre-programmed with rules, but rather learn from large amounts of data and respond to user input by analysing said data. A good guide on ML based chatbots can be found at (IBM Watson Advertising,\href{https://www.ibm.com/watson-advertising/thought-leadership/machine-learning-chatbot}{“The ultimate guide to machine-learning chatbots and conversational AI”},2022).\cite{Advertising}
Today chatbots are being used in various domains such as customer service, healthcare, and even education \cite{brandtzaeg2017people}(Brandtzaeg, 2017).  A specific common chatbot application is one where they are used to provide product support of an enterprise’ products, thus requiring less human involvement in customer support.
The foundation of many modern chatbots is the Large Language Model (LLM), a model that processes natural language input and generates words based on the data seen. Their impact and capabilities are discussed in the paper \cite{tamkin2021understanding}(Alex Tamkin,\href{https://arxiv.org/pdf/2102.02503.pdf}{“Understanding the Capabilities, Limitations, and Societal Impact of Large Language Models”},2021). 
While LLMs have made chatbots very human-like in their customer interactions, they also gave rise to a new problem known as \href{https://www.thehindu.com/sci-tech/technology/explained-what-are-hallucinating-chatbots/article66520383.ece} {Hallucination} (Nabeel Ahmed, Feb 2022). Hallucination is a phenomenon where an AI model creates results that are not real, do not match any data the algorithm has been trained on, or do not follow any other discernible pattern. It is thius important to employ a combination of evaluation techniques, to assess the presence of hallucination in large language models, especially when used in the context of chatbots. 
With the increasing popularity of chatbots, it is important to develop methods to evaluate their performance.   A chatbot benchmark is a standardized evaluation framework used to assess the performance and capabilities of chatbot systems. It involves defining a set of tasks or criteria that the chatbot must fulfil, and then measuring its performance against those tasks or criteria. There are two important factors to consider while benchmarking a chatbot: accuracy and usefulness. Chatbot accuracy is a measure of how factually correct it is, while usefulness measures to what extent the chatbot meets the user’s needs.  Not that while earlier an apparently useful answer would almost always be useful, with the deployment of hallucination-prone LLM-powered chatbots, that is no longer the case -i.e. it is possible for a chatbot to hallucinate up an answer that is not supported by any of the available documents, and can be completely wrong.  Thus to successfully evaluate the chatbot’s performance, both these factors need to be considered. 
There are various types of chatbot benchmarks, each with their own sets of advantages and disadvantages. Some examples of such benchmarks are discussed in \cite{wang2022modern}(Wang, 2022).  An example of a category of chatbot benchmarks are those that compare the answer to a question from a chatbot with that generated by a human.  While there are several other benchmarks such as Summarisation benchmarks, Information Retrieval benchmarks, and more, such benchmarks provide the best option for a user-centric evaluation, as it emphasizes the user’s perspective.
Inspired by this category, we propose the End-to-End (E2E) Chatbot Benchmark which compares the answer provided by the chatbot to an expert human answer (also known as a golden answer) using certain semantic similarity metrics that try to compare the “meanings” of each answer, as opposed to just comparing their exact words which is the current state-of-the-art (SOTA) in benchmarking.  The E2E benchmark also measures all the optimization factors previously mentioned for a successful evaluation. Overall, the E2E benchmark provides a user-centric, standardized approach and promotes real-world relevance. However, it does need a set of Golden answers. In this paper, we propose the semantic similarity aspect of the E2E benchmark to be measured using the cosine similarity measure among their corresponding text embeddings.
This paper discusses the E2E Chatbot Benchmar, and how it improves the assessment and evaluation of chatbots, compared to another well-used SOTA alternative, ROGUE scores.

\section{Related Work}

Chatbot output testing involves giving inputs to the chatbot and analyzing the output given by the chatbot.
The analysis is done to gain insight about the chatbot and to measure various relevant metrics like the relevance, completeness, accuracy, and recall.
Various metrics have been used by researchers to evaluate the performance of chatbots which include but are not limited to the use of benchmarks and metrics like GLUE, BLEU, perplexity, ROUGE score and human evaluation. 
In the subsequent section we shall explore the various metrics.
Chatbots should be accurate, relevant, and useful for them to satisfy user needs. Accuracy is indicative of how many correct answers the chatbot is able to provide, relevance is usually measured with the help of precision and recall. 
Perplexity is a commonly used intrinsic metric for evaluation of any language model, it is used to evaluate how well a model predicts text samples, that is- it is used to measure the degree of uncertainty when a model generates a new token therefore perplexity scores having lower values are indicative of a better performance by the model.
\href{https://arxiv.org/pdf/1804.07461.pdf}{Alex Wang et al. (2019)} in their conference paper published at ICLR presented another benchmark- the General Language Understanding Evaluation (GLUE) benchmark for language model evaluation which is used to evaluate the general language understanding capabilities of a language model. It consists of a collection of diverse tasks for NLU (Natural Language Understanding) along with its own datasets and evaluation metric, it attempts to provide a single framework for assessing model across various tasks however criticism about it not encompassing the entire range of tasks, lack of a diverse dataset and lack of multi-lingual assessments have been made.
\href{https://aclanthology.org/P02-1040/}{Bilingual Evaluation Understudy(BLEU)} [Kishore Papineni et al.(2002)] is another metric commonly used for evaluation of machine translation of a text from one language to another.
BLEU determines the accuracy of individual n-grams by tallying the count of matching n-grams between the candidate and reference translations. Furthermore, it takes into account the conciseness of the candidate translation relative to the references. These precision scores are harmoniously combined through a weighted geometric mean, wherein the brevity penalty plays a role, resulting in the ultimate BLEU score.
\href{https://aclanthology.org/W04-1013}{BLEU} was one of the first metrics to claim a high correlation with human judgement, and still is one of the most popular and inexpensive metrics for translation evaluation, however it is not comparable across different datasets and does not take into account intelligibility or grammatical correctness.
ROUGE or Recall Oriented Understudy for Gisting Evaluation is used for evaluating the text summarizing capabilities of the model wherein the goal is to generate a concise summary of a longer text. ROUGE score measures the similarity between the machine-generated summary and the reference summaries using overlapping n-grams, word sequences that appear in both the machine-generated summary and the reference summaries. The most common n-grams used are unigrams, bigrams, and trigrams.
ROUGE score is important as it gives flexibility to use n-grams based on requirements, however it may not fully capture semantic meaning or coherence of summary, which is important for quality of the summary.
Human evaluation is also a very important criterion that any language model or chatbot should excel in, however it can be very subjective and prone to biases, also large scale human evaluation can be very expensive and time consuming.

\section{Benchmarks in Relation to Chatbots}

This figure gives an insight into a typical chatbot designed for information retrieval from a knowledge graph to user’s natural queries that are (optionally) subsequently summarized by an LLM.  It also provides different benchmarks that can be constructed to tap into the different stages 

\begin{figure}[H]
  \includegraphics[width=\linewidth]{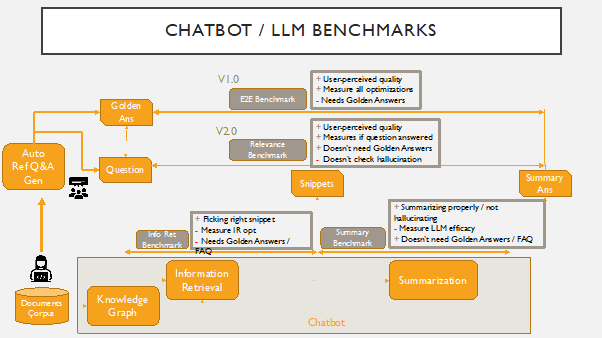}
  \caption{A typical LLM-powered chatbot for answering questions based on a document corpus and the various benchmarks that can be used to evaluate it.}
\end{figure}

The E2E benchmark uses a set of “Golden Answers” to accurately measure the chatbot’s performance. Golden Answers are a set of pre-defined answers to which the chatbot’s answers are compared. Through this process, we can evaluate the chatbot’s performance. Golden answers also come with their own set of advantages and disadvantages. These answers need to be defined before runtime (manually) and thus can prove to be tedious, time-consuming, and do not provide real-time feedback. However, several advantages make golden answers an extremely useful evaluation method such as consistency, standardized evaluation, and providing vital information for improving chatbot performance. The E2E benchmark aims to simulate real-life scenarios and assess the chatbot’s performance in handling diverse user queries. A combination of objectivity and subjectivity is created by mixing metrics such as accuracy, precision, etc., and human evaluators that provide subjective ratings and feedback. Thus, the benchmark provides an experience based around the user, as they are the ones that assess the chatbot’s performance based on criteria such as relevance, correctness, etc. Through these methods the E2E benchmark considers both factors involved in chatbot benchmarking, accuracy, and usefulness and successfully measures all optimizations. Other benchmarking methods have their advantages and disadvantages. The Relevance benchmark also creates a user-centric experience and does not require the aid of golden answers; however, it does not account for hallucination, which is the biggest drawback of Large Language Models. The Summarization benchmark does not suffer from hallucination issues and does not require golden answers, but it has a narrow focus and is not accurate in measuring the important aspects of LLM efficacy, such as language understanding, response generation, or context retention. The previous two benchmarks are also unsupervised learning methods. The Information Retrieval benchmark serves as a search benchmark rather than a chatbot benchmark. Considering all these factors, the E2E benchmark proves to be the best way to evaluate chatbot performance in this scenario. 
In this paper we are introducing an additional technique to E2E benchmarking -beyond word / n-gram based metrics like ROGUE and BLEU.  Namely we are proposing that if we measure the semantic similarity between golden answers and chatbot generated answers, they constitute a more sensitive metric to measure any improvements or degradations in the chatbot system.
Specifically in this paper we compare three different benchmarks, the first two -one using the Universal Sentence Encoder (USE) (Daniel Cer, 2018)\cite{cer2018universal} as a semantic embedding, another using Sentence Transformers (ST) (Nils Reimers, 2019)\cite{reimers2019sentence} as a semantic embedding, and we compare them with the various components of ROGUE as a legacy word or n-gram based metric.
To conduct the analysis, three libraries have been used, namely:\\
1.	Universal Sentence Encoder\\
2.	Sentence Transformer\\
3.	ROUGE (Recall-Oriented Understudy for Gisting Evaluation)\\
As previously discussed, this paper will primarily be focusing on the E2E benchmark as compared to other methods.  In the following sections of the paper, the results of this evaluation will be discussed.\\

\section{Results}
We propose here a semantic-similarity based End-to-End (E2E) Benchmark for benchmarking performance of chatbots.  This benchmarking compares and checks the answers from the chatbot against a ‘golden answer’ provided by a human expert using a semantic similarity measure based on text embeddings.  We also compare the result we get from the such-constructed E2E Benchmark vs a more traditional approach of comparing the word, bi-gram, and n-gram based ROGUE scores computed on the same pairs of ‘golden answer’ vs chatbot answers.\\
We used three libraries to conduct our analyses which included\\
1. Universal Sentence Encoder\\
2. Sentence Transformer\\
3. ROUGE\\
Before continuing the analysis, we would like to give a brief idea about \href{https://arxiv.org/abs/1803.11175}{Universal Sentence Encoder} and \href{https://arxiv.org/abs/1908.10084}{Sentence Transformers}.

\begin{figure}[H]
    \centering
    \includegraphics[width=\linewidth]{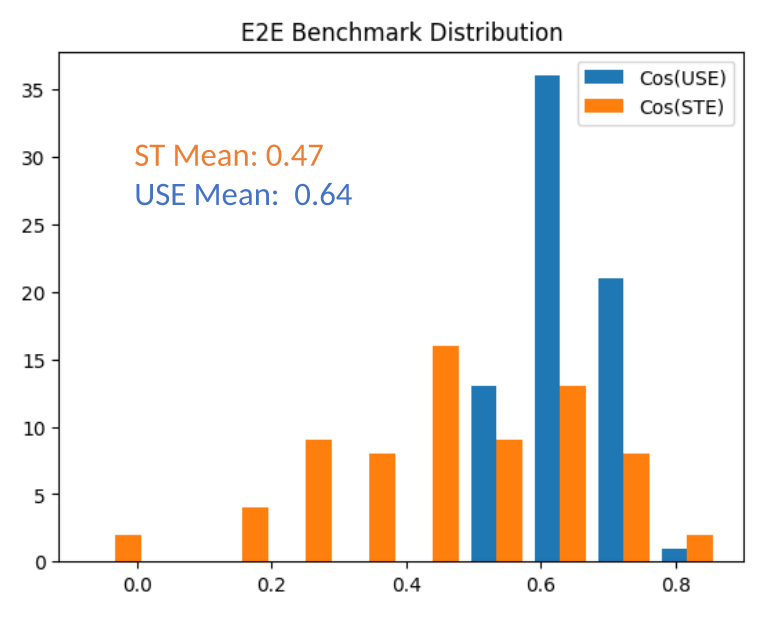}
    \caption{E2E Benchmark results with use of Cosine Similarity over embeddings obtained from the Golden Answer vs the Chatbot generated Answer -using both Sentence Transformer (ST) and Universal Sentence Encoder (USE) embeddings, respectively.  Result obtained from a product support chatbot.}
\end{figure}

Universal Sentence Encoder (USE) is a pre-trained model developed by Google that converts sentences into fixed-length vector representations, or embeddings, capturing their semantic meanings.

The Sentence Transformer (ST) is a model architecture designed to produce high-quality sentence embeddings, which represent sentences as vectors. Unlike traditional word embeddings, these sentence embeddings consider the complete context and meaning of the sentences. Drawing inspiration from transformer architectures like BERT, the Sentence Transformer model utilizes pre-training and fine-tuning techniques.

For our purposes, we evaluated the performance of a publicly available product support chatbot made by a large network equipment company to support their enterprise users get answers to questions based on their documentation.

For both the embeddings, we take a golden answer (G) and the chatbot generated answer (A), the we apply the embedding \(X=F(T)\) where F can be either a Universal Sentence Encoder or a Sentence Transformer on our text T- which gives us \(X_G\) and \(X_A\) both being n-dimensional vectors generated as a result of applying the operations.  Then we calculate the cosine similarities between \(X_G\)  and \(X_A\)    .  Cosine similarity is defined as dot product of two vectors, divided by the product of their magnitudes.
$$
S_{(G, A)}=\frac{X_G \cdot X_A}{\left|X_G\right|\left|X_A\right|}
$$
In case of using USE as the embedding operation, there is usually a constant bias of 0.5 remaining in our final answer which is due to the inherent characteristics of USE vectors. 
We computed the cosine similarities for both Universal Sentence Encoder and Sentence Transformer, and for ROUGE we use precision and recall to check for words and n-grams included in our generated answers while comparing against the expert human answers.
As shown in the Figure 2, the chatbot performs reasonably well for both Universal Sentence Encoder and Sentence Transformer, with the USE mean of about 0.64 and the ST mean of about 0.47, we also calculated the correlation between the outputs of both the methods which is displayed in Figure 3.
\begin{figure}[H]
    \centering
    \includegraphics[width=\linewidth]{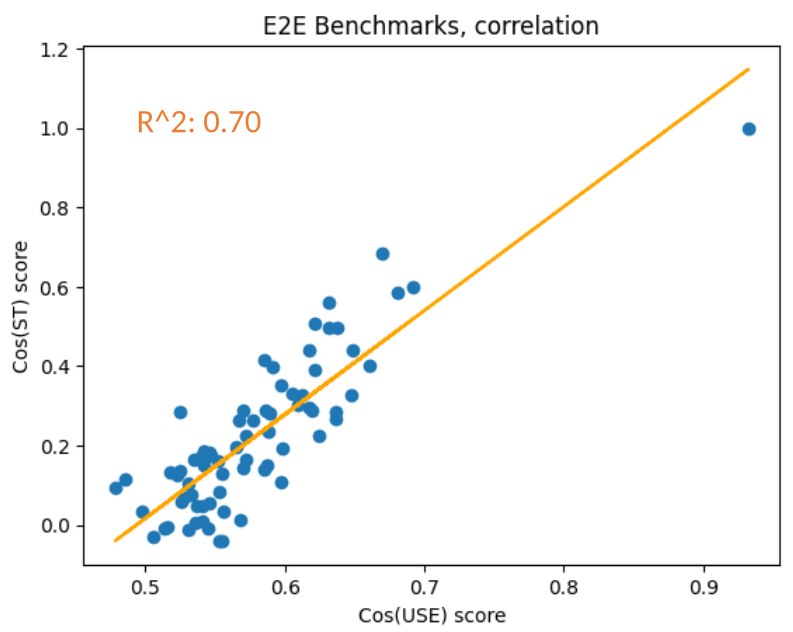}
    \caption{Correlation between E2E Benchmark results with use of Cosine Similarity over Sentence Transformer (ST) embeddings, and Universal Sentence Encoder (USE) embeddings from the same product support chatbot}
\end{figure}

We calculate the \(R^2\) scores between the cosine similarity of both the Sentence Transformer and Universal Sentence Encoder to gauge the similarity between the models for our chatbot which gives us an impressive 0.7 which indicates that the models both work pretty similar and have very similar answers to our provided expert answers.
Next, we turn our attention to ROUGE scores.
ROUGE-1 or word-level ROUGE score which evaluates the quality of answer based on the overlapping of unigrams between the answers provided by the chatbot, ROUGE-1 is useful as it captures the keywords however it fails to take into account the order or structure of the words.
We can also use the metric (1-precision) to get an estimate of the hallucination the chatbot has.
The chart below provides the ROUGE-1 scores when we tested our chatbot answers against the expert answer:
\begin{figure}[H]
    \centering
    \includegraphics[width=\linewidth]{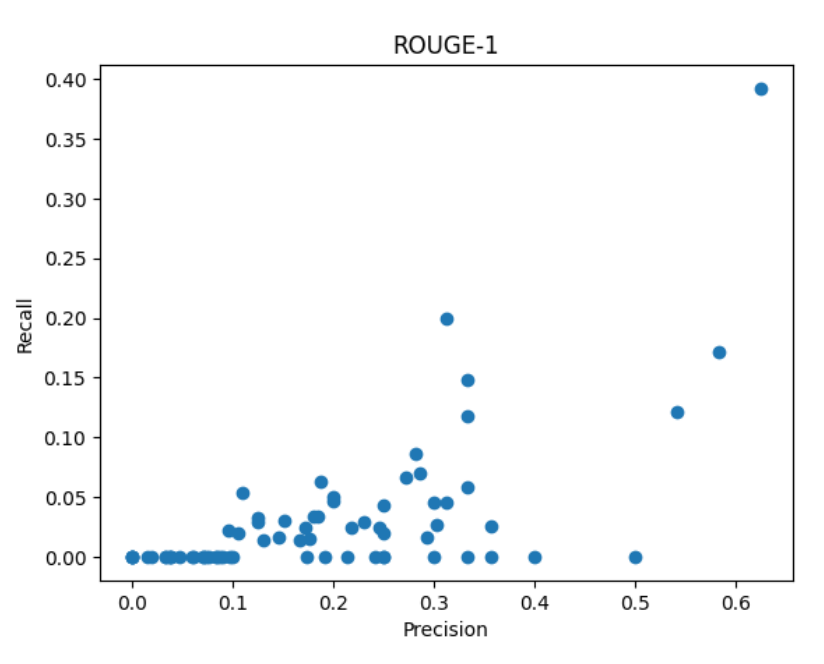}
    \caption{Precision vs Recall for ROGUE-1 scores between Golden Answers vs Chatbot Answers from the same product support chatbot}
\end{figure}

The results are very interesting.  They don’t show any direct mapping between recall vs precision at unigram level however we also can use other forms of ROUGE to get a better understanding of the model-
ROUGE-2 (bigrams) and ROUGE-LCS (longest common sequence)-  the recall vs precision of which are given in Figure 5 and Figure 6, respectively.
\begin{figure}[H]
    \centering
    \includegraphics[width=\linewidth]{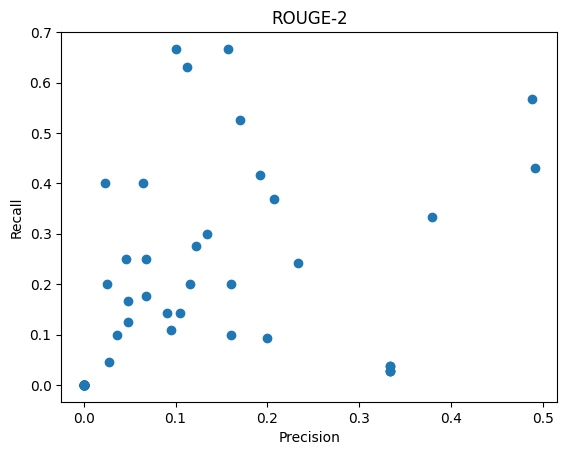}
    \caption{Precision vs Recall for ROGUE-2 scores between Golden Answers vs Chatbot Answers from the same product support chatbot}
\end{figure}

\begin{figure}[H]
    \centering
    \includegraphics[width=\linewidth]{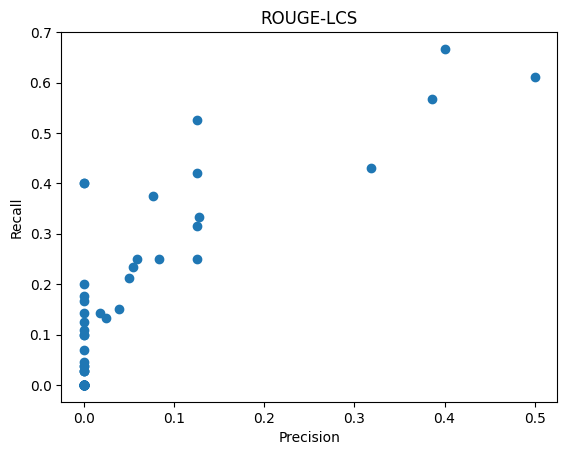}
    \caption{Precision vs Recall for ROGUE-1 scores between Golden Answers vs Chatbot Answers from the same product support chatbot}
\end{figure}

For bigrams, we see most of recall improve compared to unigram level, precision however was better when we used ROUGE-1 as it only captures one word at a time.
\section{Golden Answers vs. Random Words : A Negative Example}
We also calculated the cosine similarity as described in the above sections- this time between the golden answers and random words in order to better judge how well our experiments perform in relation to a random sample, we found out that the metrics are completely uncorrelated, with the R2 score being 0, however we also notice that Sentence Transformer outperforms USE here as USE still gets a cosine similarity of about 0.5 whereas Sentence Transformers give a cosine similarity of 0.
Image below shows the cosine scores calculated between golden answers and random words.
\begin{figure}[H]
    \centering
    \includegraphics[width=\linewidth]{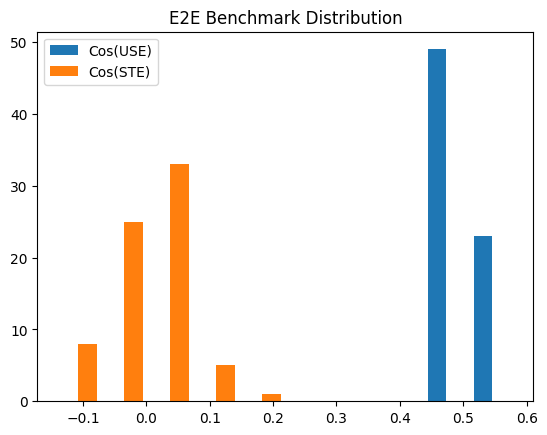}
    \caption{E2E Benchmark results with use of Cosine Similarity over embeddings obtained from the Golden Answer vs random words -using both Sentence Transformer (ST) and Universal Sentence Encoder (USE) embeddings, respectively.}
\end{figure}

We even calculated the correlation between the metrics as mentioned above the results of which are in Figure 7.
\begin{figure}[H]
    \centering
    \includegraphics[width=\linewidth]{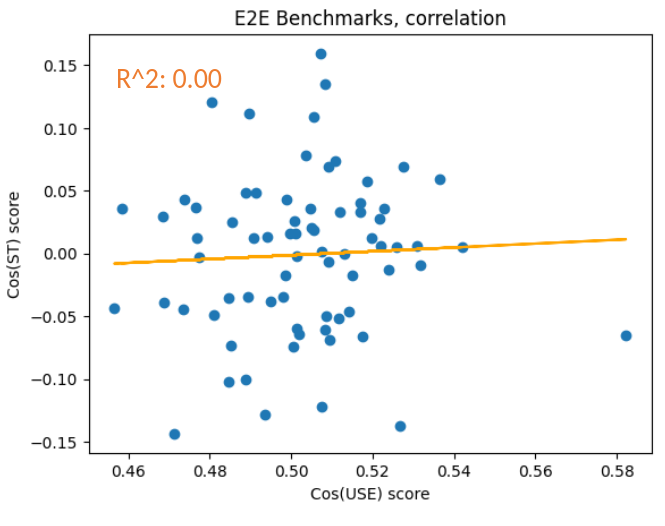}
    \caption{Correlation between E2E Benchmark results with use of Cosine Similarity over Sentence Transformer (ST) embeddings, and Universal Sentence Encoder (USE) embeddings comparing Golden Answers vs random words}
\end{figure}

As we expected, there is no correlation between the metrics in this case.
\section{Measuring improvement in the chatbot: GPT-3 with prompt engineering}
It has been observed in the literature that LLM-based chatbots can be improved using enhanced prompts, what is often called prompt engineering
The chatbot we used for our experiments uses GPT-3, and as such is sensitive to prompt engineering.  So we compare the E2E Benchmark results obtained using standard prompts vs the results obtained after using enhanced prompts- using the methods described :\\
1.) Standard prompt : \textbf{‘from this text: {}, can you answer the question: {}’}\\
2.) Enhanced prompt: \textbf{{"role": "system", "content": "You are a network engineering expert helping summarize the following message to help a question from the network operator which follows the message"}}\\
The E2E results from these experiments are in Figures 9 and 10 respectively.
\begin{figure}[H]
    \centering
    \includegraphics[width=\linewidth]{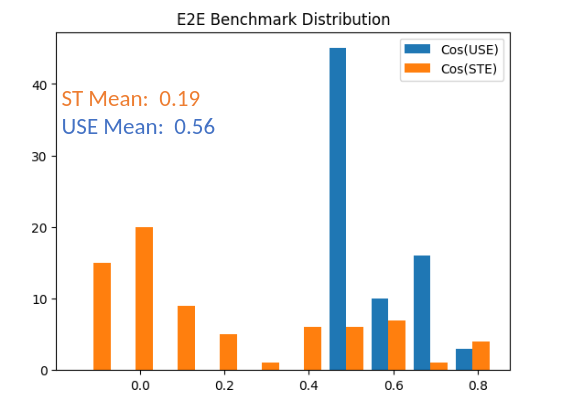}
    \caption{E2E Benchmark results with use of Cosine Similarity over embeddings obtained from the Golden Answer vs random words -using both Sentence Transformer (ST) and Universal Sentence Encoder (USE) embeddings, respectively, using the Standard Prompt}
\end{figure}

\begin{figure}[H]
    \centering
    \includegraphics[width=\linewidth]{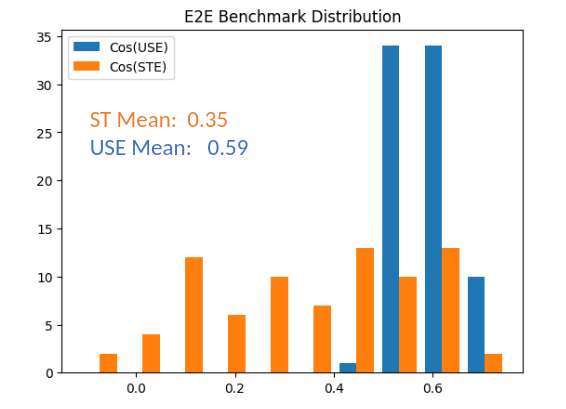}
    \caption{E2E Benchmark results with use of Cosine Similarity over embeddings obtained from the Golden Answer vs random words -using both Sentence Transformer (ST) and Universal Sentence Encoder (USE) embeddings, respectively, using the Enhanced Prompt.}
\end{figure}

We see a significant rise in Sentence Transformer outputs on engineered prompts, suggesting it is much more sensitive to prompt engineering tweaks on the query.
We also checked the performance of the chatbot against ROUGE metrics, with Standard prompts, vs Enhanced prompts.  The ROGUE-1 score results from these experiments are in Figures 11 and 12 respectively, while the ROGUE-2 score results from these experiments are in Figures 13 and 14 respectively.
\begin{figure}[H]
    \centering
    \includegraphics[width=\linewidth]{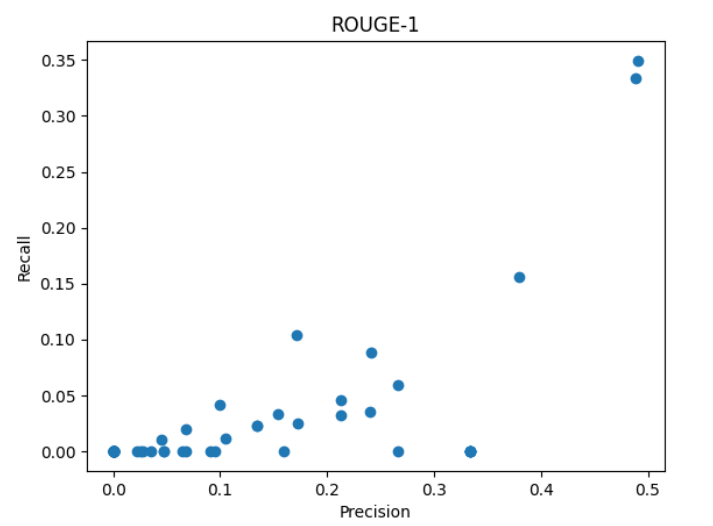}
    \caption{ROGUE-1 score comparisons between the Golden Answer vs Chatbot Answers, using the Standard Prompt.}
\end{figure}

\begin{figure}[H]
    \centering
    \includegraphics[width=\linewidth]{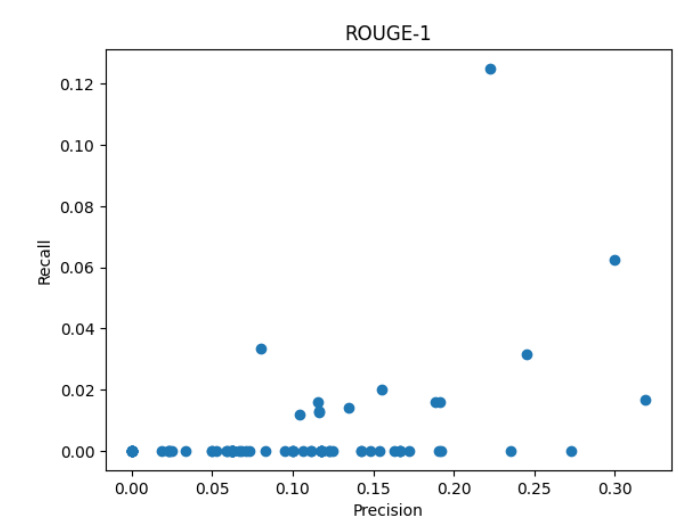}
    \caption{ROGUE-1 score comparisons between the Golden Answer vs Chatbot Answers, using the Standard Prompt. }
\end{figure}

We see that for most outputs, the improvement in precision scores do not correlate with increase in recall however ROUGE-2 capture this trend better as we see in Figures 13 and 14.
\begin{figure}[H]
    \centering
    \includegraphics[width=\linewidth]{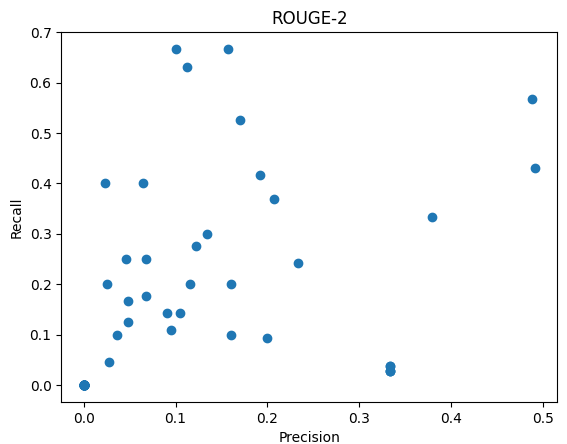}
    \caption{ROGUE-2 score comparisons between the Golden Answer vs Chatbot Answers, using the Standard Prompt.}
\end{figure}
When given engineered prompts- ROUGE-2 shows rather interesting output as shown below-
\begin{figure}[H]
    \centering
    \includegraphics[width=\linewidth]{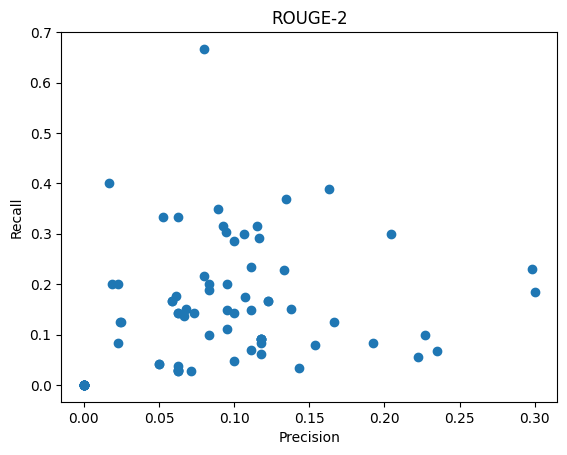}
    \caption{ROGUE-2 score comparisons between the Golden Answer vs Chatbot Answers, using the Standard Prompt.}
\end{figure}

ROUGE-2 shows clustered outputs, however the maximum values and the number of outputs close to them seems to dip for both the axes.
ROUGE scores do not show improvement on applying prompt engineering.  However the E2E benchmark metrics, i.e. cosine similarity of the USE and ST embeddings show considerable improvement on using engineered prompts.
\section{Conclusion}
Over the past few years, there has been tremendous growth in the number and quality of chatbots. To ensure that these chatbots are up to the mark we need to be able to effectively assess their performance. This has created the need for efficient chatbot benchmarking methods. Throughout this paper, many methods of chatbot benchmarking have been discussed, highlighting their various advantages and disadvantages. In addition we have proposed a new benchmark that we cal the E2E (End-to-End) Benchmark.  The E2E Benchmark is constructed using cosine similarity between sentence embeddings of the chatbot’s answers to questions vs the corresponding “golden answers” to the same questions generated by human experts.  For generating such embeddings we considered two different embedding libraries, the Universal Sentence Encoder (USE) and the Sentence Transformer (ST).
While the E2E Benchmark using USE provides a higher mean score than the same using ST, the chatbot performed well for both libraries and through the plotting their scores against each other, we observed a similarity in their scores. Replacing the golden answers by a set of random words should have resulted in a mean score of near 0 which was observed in the case of ST, but USE showed a mean score of 0.5 which was likely due to its systemic bias.  By providing enhanced prompts to the chatbot, we observed an increase in mean score for both libraries, however the ST score showed a sharper increase, indicating high sensitivity to the enhanced prompt. The ROGUE scores were also computed using two different metrics; precision which tells how concise the answer is and recall, which tells us how accurate the answer is when compared to the Golden answer. Different ROGUE scores are produced when run on different n-gram levels (e.g., unigram-ROGUE-1). The ROGUE-1 score did not show any obvious correlation between precision and recall. For ROGUE-2, we observed an improvement in recall however, the precision score was not as good compared to ROGUE-1. Applying Prompt Engineering on the LLM part of the chatbot, we did not significantly improve the ROGUE scores -while the E2E Benchmark did show significant improvement.  Through our analysis, we concluded that the E2E Benchmark, when it is used with the ST library, works best as it performed well in all scenarios and showed significant improvements at each stage. E2E Benchmarks using USE followed that as the second best.  While the ROGUE scores were too unpredictable, and as a result, we can’t recommend them as a reliable means to benchmark chatbots’ performance. In summary, while chatbot benchmarking remains an open area of research, after careful analysis and applying different methods to evaluate performance we firmly believe that the use of the E2E Benchmark as we proposed, is the best option.
\bibliographystyle{ieeetr}
\bibliography{citations}

\begin{thebibliography}{1}

\bibitem{Advertising}
I.~W. Advertising, ``The ultimate guide to machine-learning chatbots and
  conversational ai: Ibm watson advertising.''

\bibitem{brandtzaeg2017people}
P.~B. Brandtzaeg and A.~F{\o}lstad, ``Why people use chatbots,'' in {\em
  Internet Science: 4th International Conference, INSCI 2017, Thessaloniki,
  Greece, November 22-24, 2017, Proceedings 4}, pp.~377--392, Springer, 2017.

\bibitem{tamkin2021understanding}
A.~Tamkin, M.~Brundage, J.~Clark, and D.~Ganguli, ``Understanding the
  capabilities, limitations, and societal impact of large language models,''
  {\em arXiv preprint arXiv:2102.02503}, 2021.

\bibitem{wang2022modern}
Z.~Wang, ``Modern question answering datasets and benchmarks: A survey,'' {\em
  arXiv preprint arXiv:2206.15030}, 2022.

\bibitem{cer2018universal}
D.~Cer, Y.~Yang, S.-y. Kong, N.~Hua, N.~Limtiaco, R.~S. John, N.~Constant,
  M.~Guajardo-Cespedes, S.~Yuan, C.~Tar, {\em et~al.}, ``Universal sentence
  encoder,'' {\em arXiv preprint arXiv:1803.11175}, 2018.

\bibitem{reimers2019sentence}
N.~Reimers and I.~Gurevych, ``Sentence-bert: Sentence embeddings using siamese
  bert-networks,'' {\em arXiv preprint arXiv:1908.10084}, 2019.

\end{thebibliography}
\end{document}